\documentclass[10pt,twocolumn,letterpaper]{article}

\usepackage{cvpr}
\usepackage{times}
\usepackage{epsfig}
\usepackage{graphicx}
\usepackage{amssymb}
\usepackage{multicol}
\usepackage{multirow}
\usepackage{stfloats}
\usepackage{amsthm}
\usepackage{empheq}
\usepackage{amsmath}
\DeclareMathOperator{\sign}{sign}

\usepackage[pagebackref=true,breaklinks=true,letterpaper=true,colorlinks,bookmarks=false]{hyperref}

\graphicspath{{./images/}}

\cvprfinalcopy %

\def\cvprPaperID{13} %

\ifcvprfinal\fi
\begin{document}

\title{Scene Motion Decomposition for Learnable Visual Odometry}
\author{Igor Slinko, Anna Vorontsova, Filipp Konokhov, Olga Barinova, Anton Konushin \\
Samsung AI Center Russia, Moscow\\
{\tt\small \{i.slynko,a.vorontsova,f.konokhov,o.barinova,a.konushin\}@samsung.com}\\
}
\maketitle

\begin{abstract}
   Optical Flow (OF) and depth are commonly used for visual odometry since they provide sufficient information about camera ego-motion in a rigid scene. We reformulate the problem of ego-motion estimation as a problem of motion estimation of a 3D-scene with respect to a static camera. The entire scene motion can be represented as a combination of motions of its visible points. Using OF and depth we estimate a motion of each point in terms of 6DoF and represent results in the form of motion maps, each one addressing single degree of freedom. In this work we provide motion maps as inputs to a deep neural network that predicts 6DoF of scene motion. Through our evaluation on outdoor and indoor datasets we show that utilizing motion maps leads to accuracy improvement in comparison with naive stacking of depth and OF. Another contribution of our work is a novel network architecture that efficiently exploits motion maps and outperforms learnable RGB/RGB-D baselines.
\end{abstract}

\section{Introduction}

Visual odometry (VO) is an essential component of simultaneous localization and mapping (SLAM) systems used in robotics, 3D scanning and AR/VR applications. The goal of visual odometry is to estimate camera motion between two frames, which may be RGB/RGB-D images or other modalities. A wide range of classical geometry-based visual odometry algorithms for RGB and RGB-D have been proposed in the last few years \cite{engel2018direct, forster2014svo}. Although geometry-based algorithms show state-of-the-art accuracy on public benchmarks, they usually require a lot of manual fine-tuning to work in new environments. Excellent performance of deep learning-based methods for OF estimation \cite{dosovitskiy2015flownet, ilg2017flownet, sun2018pwc} led to emergence of end-to-end VO algorithms utilizing OF \cite{almalioglu2018ganvo, costante2018ls, wang2017deepvo}. These works demonstrate competitive accuracy on public benchmarks while having much fewer parameters compared to classical geometric methods.

\begin{figure}[t]
  \begin{centering}
    \begin{tabular}{c}
     Dominant motion: translation along the $x$ and $y$ axes \\
     \includegraphics[width=1.0\linewidth]{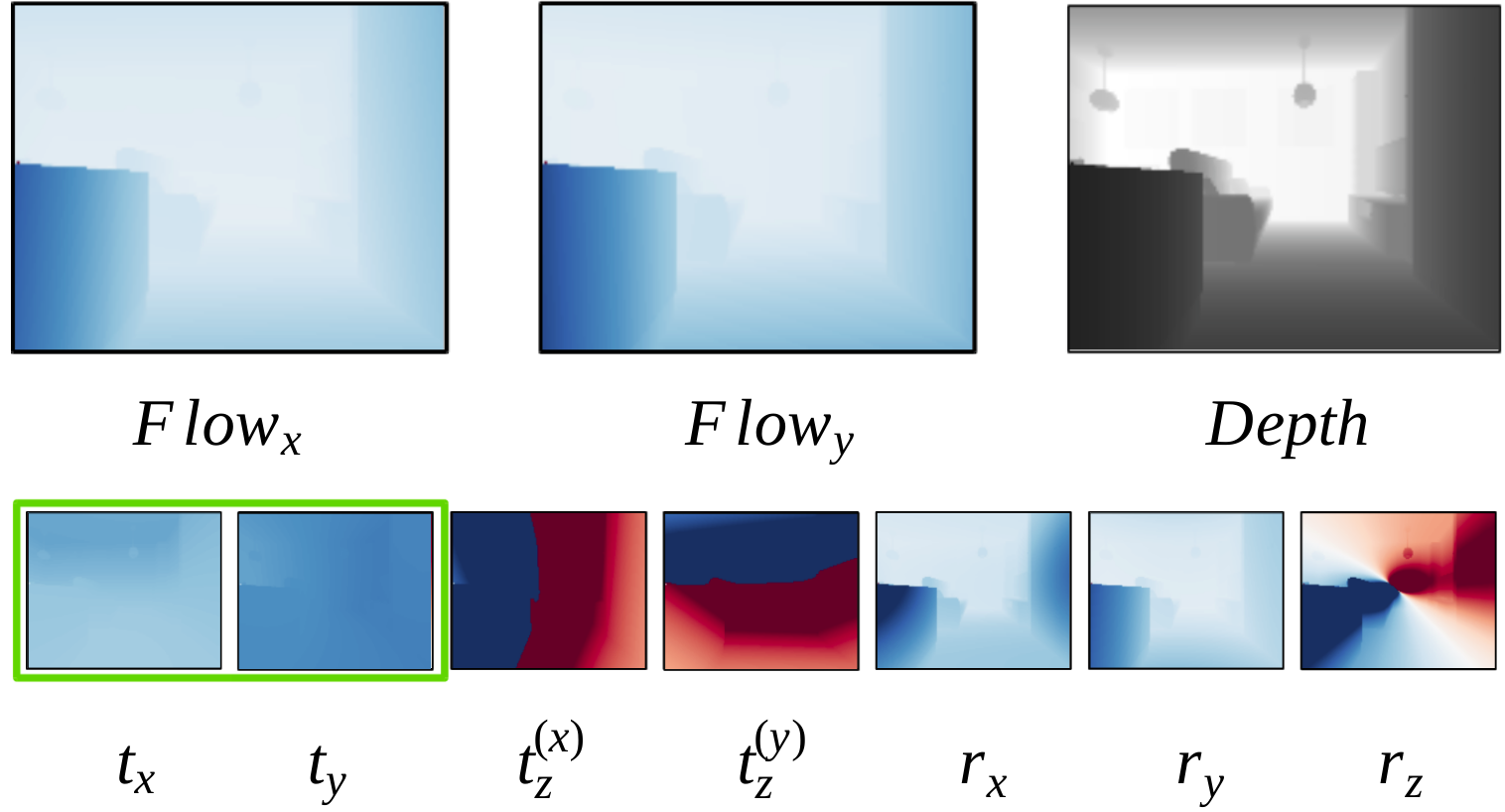}
     \label{fig:main_idea_tx_ty} \\
     \\
     Dominant motion: rotation around the $x$ and $y$ axes \\
     \includegraphics[width=1.0\linewidth]{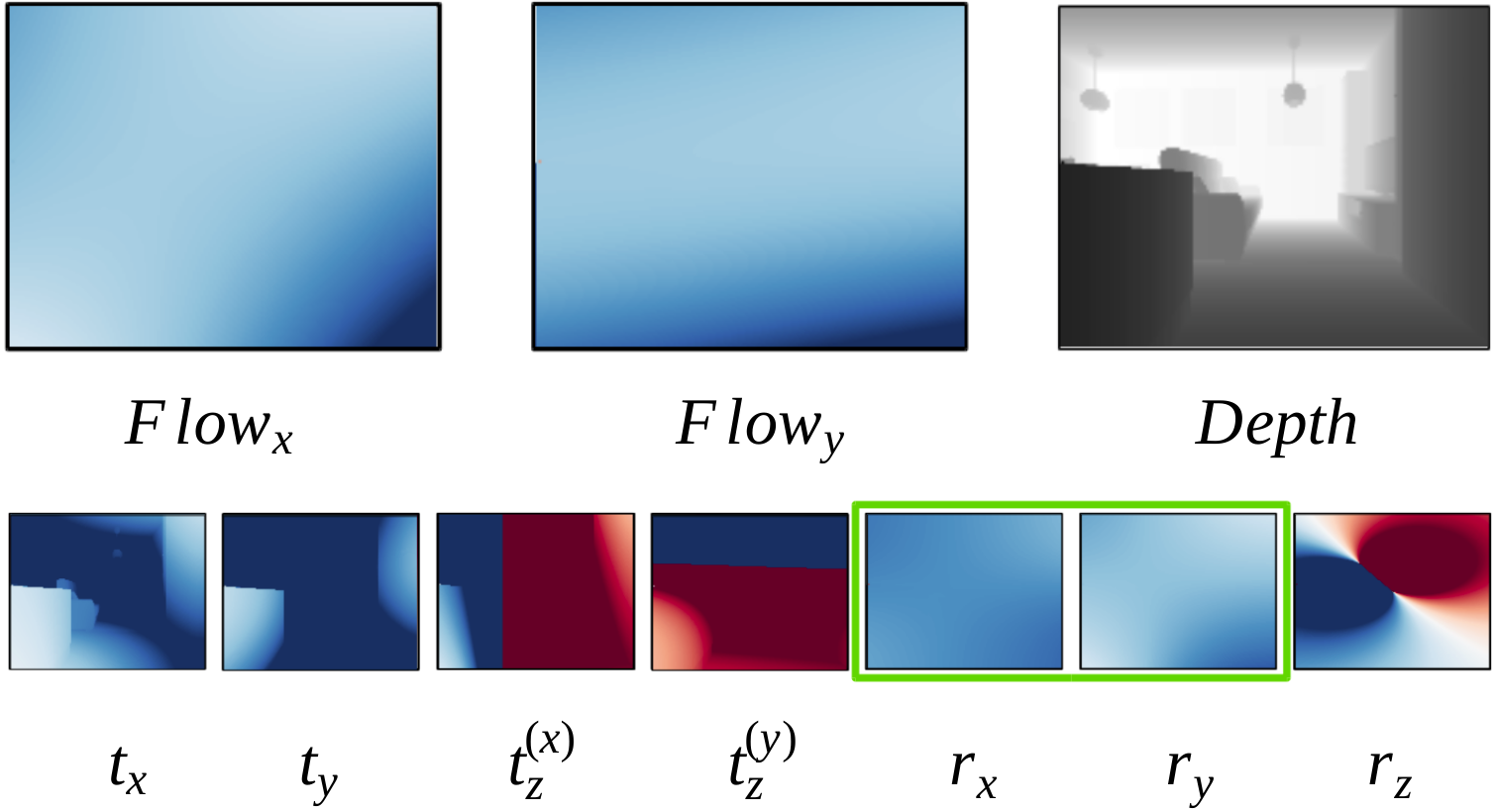}
     \label{fig:main_idea_rx_ry}
     \end{tabular}
  \end{centering}
\caption{ 
Scene motion decomposition into 6 motion maps, corresponding to 6DoF of camera motion.
In the examples above the dominant camera motion is a combination of two translations (top) or two rotations (bottom), while the other components of the motion have much lower magnitude. The resulting $t_x$, $t_y$ motion maps (top) and $r_x$, $r_y$ motion maps (bottom) are nearly constant and provide a tight approximation for a corresponding 6DoF value. For first picture, average map values are 0.3 and 0.2, while groundtruth $r_x$, $r_y$ equal to 0.3 and 0.21 respectively. For the second example, average values of motion maps corresponding to $t_x$ and $t_y$ maps are equal to 0.2, 0.3 and groundtruth values are 0.21 and 0.31.  
}
\end{figure}

Depth sensors have made great progress over the last few years and now they are accurate, cheap, and miniature enough to fit in the front panel of a cellphone. At the same time stereo and monocular depth reconstruction \cite{casser2019struct2depth, fu2018deep, liang2018learning, luo2016efficient} now reaches the accuracy of the depth sensors on public benchmarks and work in real-time \cite{icra_2019_fastdepth} or even on-a-chip \cite{Ambarella}. OF estimation is also becoming a mature technology with production-ready solutions available on the market \cite{nvidia_of_sdk}.

Recent works on deep visual odometry use both OF and depth. \cite{dharmasiri2018eng, ummenhofer2017demon}. The mentioned methods estimate motion while performing refinement of depth and flow in a single iterative pipeline. Extension to multi-task learning increases the amount of computation and makes these methods impractical for real-time applications. Moreover, in these works OF, depth and other data (such as image pair, OF confidence, first frame warped) are stacked in a simple way that may be not the best solution, since the inputs belong to different domains. 

\begin{figure*}[t]
    \begin{center}
        \includegraphics[width=1\linewidth,scale=2]{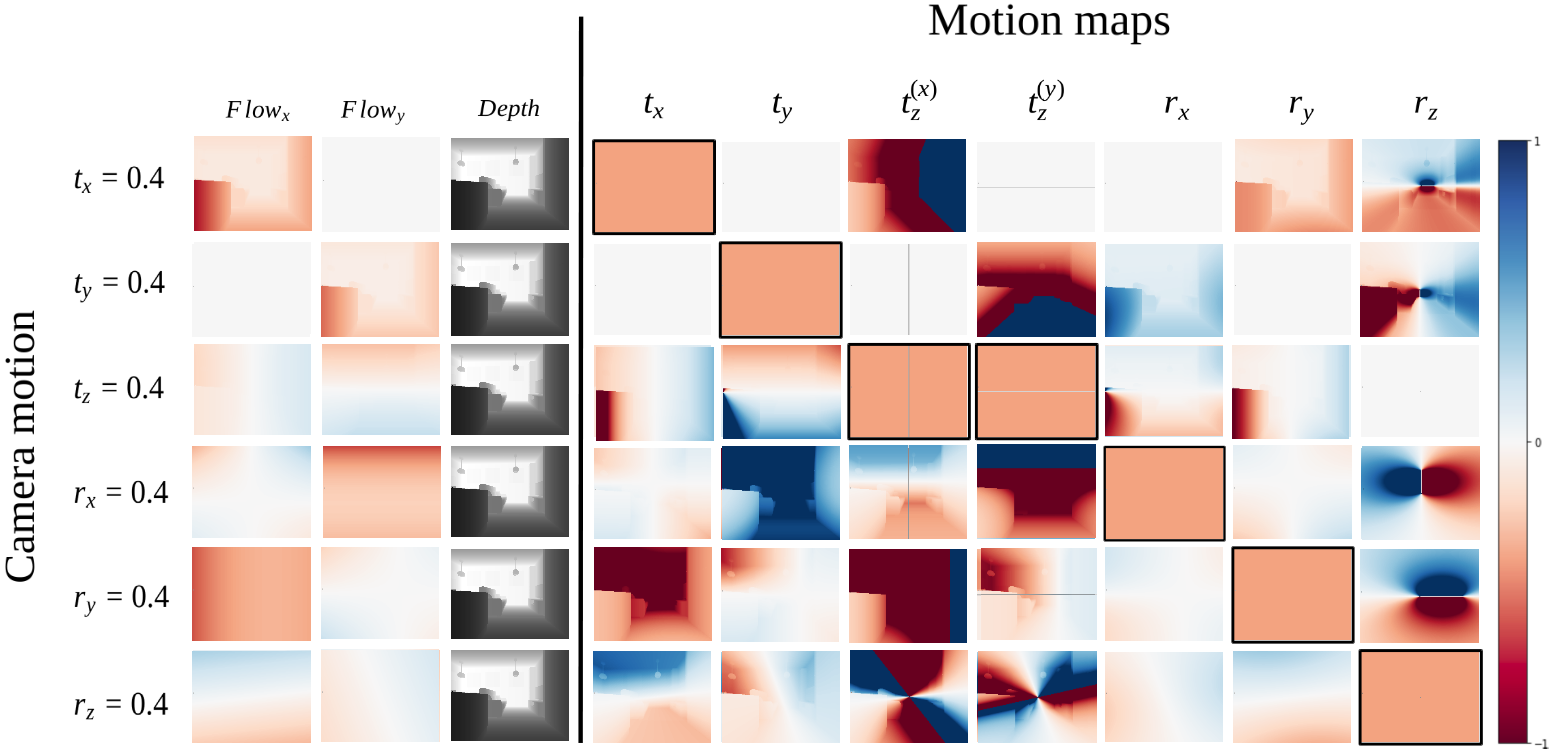}
        \caption{With only one component of 6DoF changing, motion maps provide an easily interpretable representation of the camera motion. Moreover, motions maps altogether illustrate inter-dependencies of 6DoF components.}
        \label{fig:motion_maps}
    \end{center}
\end{figure*}

In this work we focus on the novel ways of using depth and OF for deep visual odometry. We reformulate the problem of camera motion estimation as a problem of motion estimation of a 3D-scene with respect to a static camera. Our main contribution is a decomposition of scene motion obtained from OF and depth into 6 components, each one corresponding to a single degree of freedom. These components, which we call \textit{motion maps}, are based on projective geometry and require minimum computation. At the same time, passing them as inputs to the network improves quality over using OF stacked with depth.

Another contribution of our work is a novel network architecture for visual odometry that can efficiently process different inputs: OF, OF stacked with disparity (inverse depth) or motion maps. Since some regions in input maps (OF or disparity) may not be completely accurate we provide the network with attention mechanism by employing gated convolutions\cite{lei2015semi}.

We evaluate our proposed method on KITTI \cite{Geiger2012CVPR} and DISCOMAN (in submission, attached) datasets and show improved results over trainable monocular RGB and RGB-D baselines. In addition, we present comparison with classical methods, where our method shows significantly higher robustness and competitive accuracy even without use of global optimization.

\section{Related work}
\subsection{Classical methods}
Several different mathematical formulations for visual odometry have been considered in the literature. Geometry-based visual odometry methods can be  classified into direct (\eg~\cite{kerl2013robust}) or indirect (\eg~\cite{mur2017orb}) and dense (\eg~\cite{steinbrucker2011real}) or sparse (\eg~\cite{engel2018direct}). Many of the classical works use bundle adjustment or pose graph optimization for the last several keyframes in order to mitigate the odometry drift. In this work we focus on ego-motion estimation and do not perform any global pose estimation, which may further improve the performance.

\subsection{Learning-based methods} 
DeepVO \cite{wang2017deepvo} was a pioneer work to use deep learning for visual odometry. Their deep recurrent network regresses camera motion using pre-trained FlowNet \cite{dosovitskiy2015flownet} as a feature extractor. ESP-VO \cite{wang2018end} extends this model with sequence-to-sequence learning and introduces an additional loss on global poses. LS-VO \cite{costante2018ls} also uses the result of FlowNet and formulates the problem of egomotion estimation as finding a low-dimensional subspace of the OF space. DeMoN \cite{ummenhofer2017demon} estimates both egomotion, OF and depth in the EM-like iterative network. By efficient usage of two frames they improve accuracy of depth prediction over single-view methods. ENG \cite{dharmasiri2018eng} with quite close ideas achieves fair performance on both indoor and outdoor datasets.

In contrast to most works on learnable visual odometry which rely on the network to infer geometric constraints from OF and depth, we explicitly use projective geometry to compute the \textit{motion maps}. The closest work to ours is \cite{zhao2018learning}, where depth and OF are combined to generate depth flow, which is in some sense similar to our $t_{z}$ \textit{motion map}. However, using OF along with depth flow is debatable, as they have different physical meanings: first one represents pixel-shift in image-plane, second represents z-motion for each visible point in the actual scene. In our experiments we show that the use of motion maps leads to improved accuracy compared to depth flow \cite{zhao2018learning}.

\section{Motion maps}

\subsection{Motivation behind scene motion decomposition}

Optical flow provides pixel shifts in an image plane. However, such representation may not be perfect, due to several reasons described below:
\setlength{\leftmargini}{10pt}
\begin{itemize}
 \setlength{\parskip}{0pt} 
    \item[--] OF does not provide any information about point shifts along z-axis.
    \item[--] While moving along x or y axis, the value of particular pixel shift depends on distance to corresponding physical point.
    \item[--] For rotations and translation in z-axis, the value of pixel shift depends on its coordinates on image plane, that might provide an additional difficulty for convolutional neural networks.
    \item[--] Given a small fragment of OF map, it is not possible for the network to distinguish between translation and rotation camera motions.
\end{itemize}

We can reformulate the problem of ego-motion estimation as a problem of estimation motion of 3D-scene with respect to a static camera. We propose a scene motion decomposition, by which we eliminate perspective transformations.

Suppose that the scene is rigid and all but one 6DoF components of a scene motion are zero. Given OF and a depth map from the first frame, we can estimate the value of each of 6DoF motion component for each point individually.

Proposed decomposition can be especially useful for deep visual odometry methods, since it allows a network to focus on regression task instead of learning to perform complicated geometrical transforms. Below we explain how to obtain the scene motion decomposition from OF and depth.

\subsection{Calculation of motion maps}
\label{subsec:calculation_of_motion_maps}

\begin{table*}[!hbt]
  \begin{center}
    \begin{tabular}{cccc}
        \includegraphics[width=0.24\linewidth,scale=2]{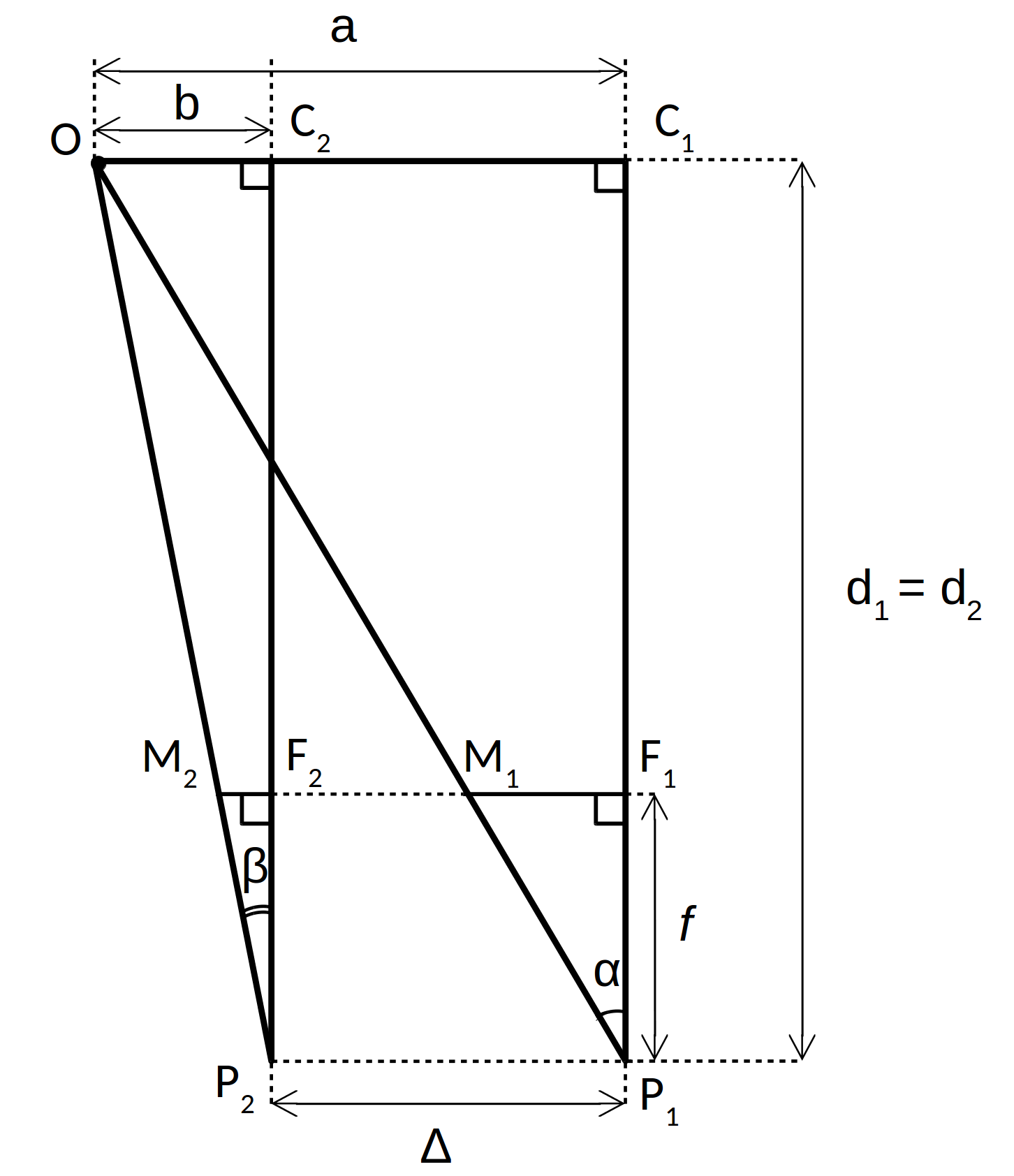}
        & \includegraphics[width=0.24\linewidth,scale=2]{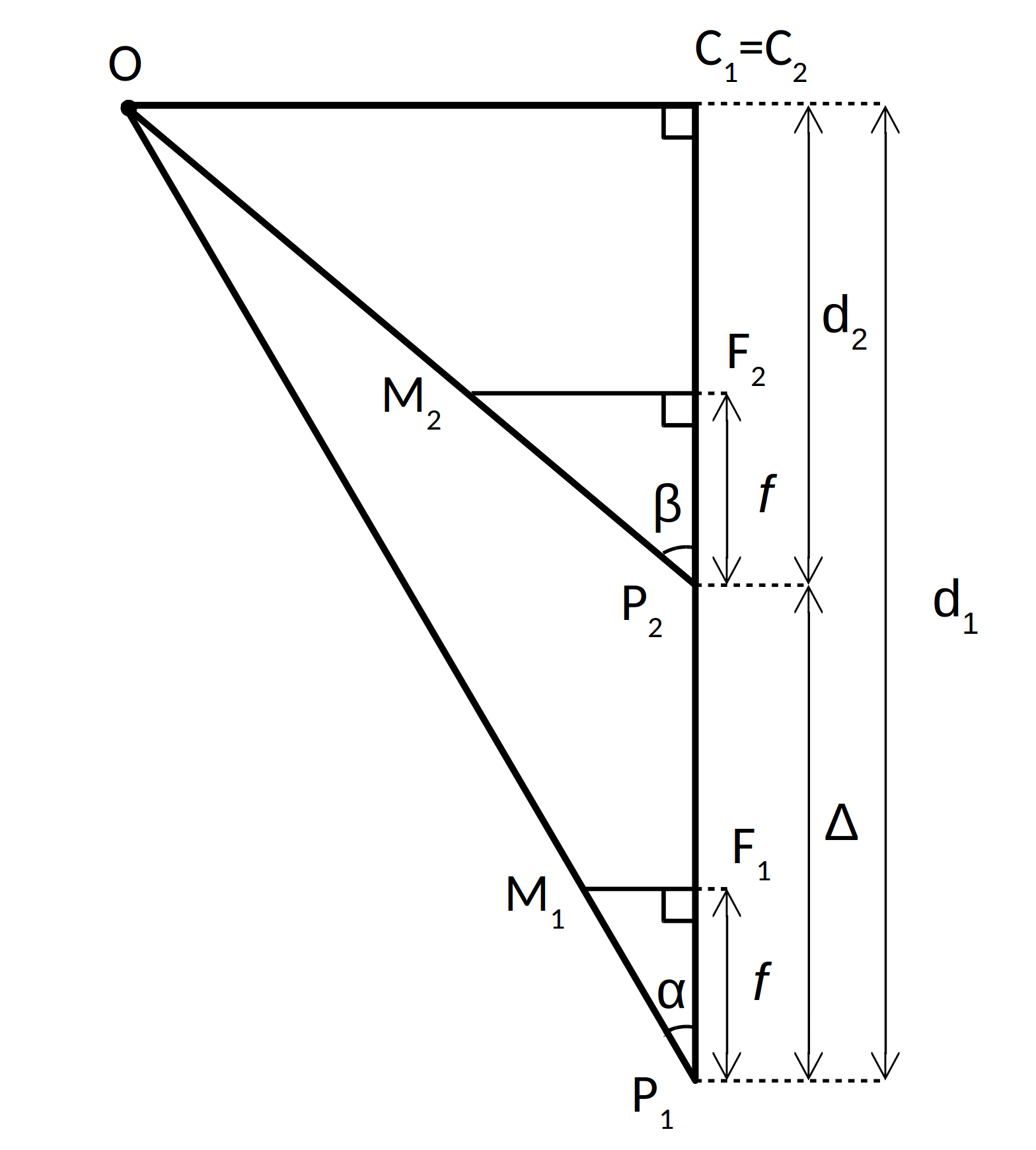}
        & \includegraphics[width=0.25\linewidth,scale=2]{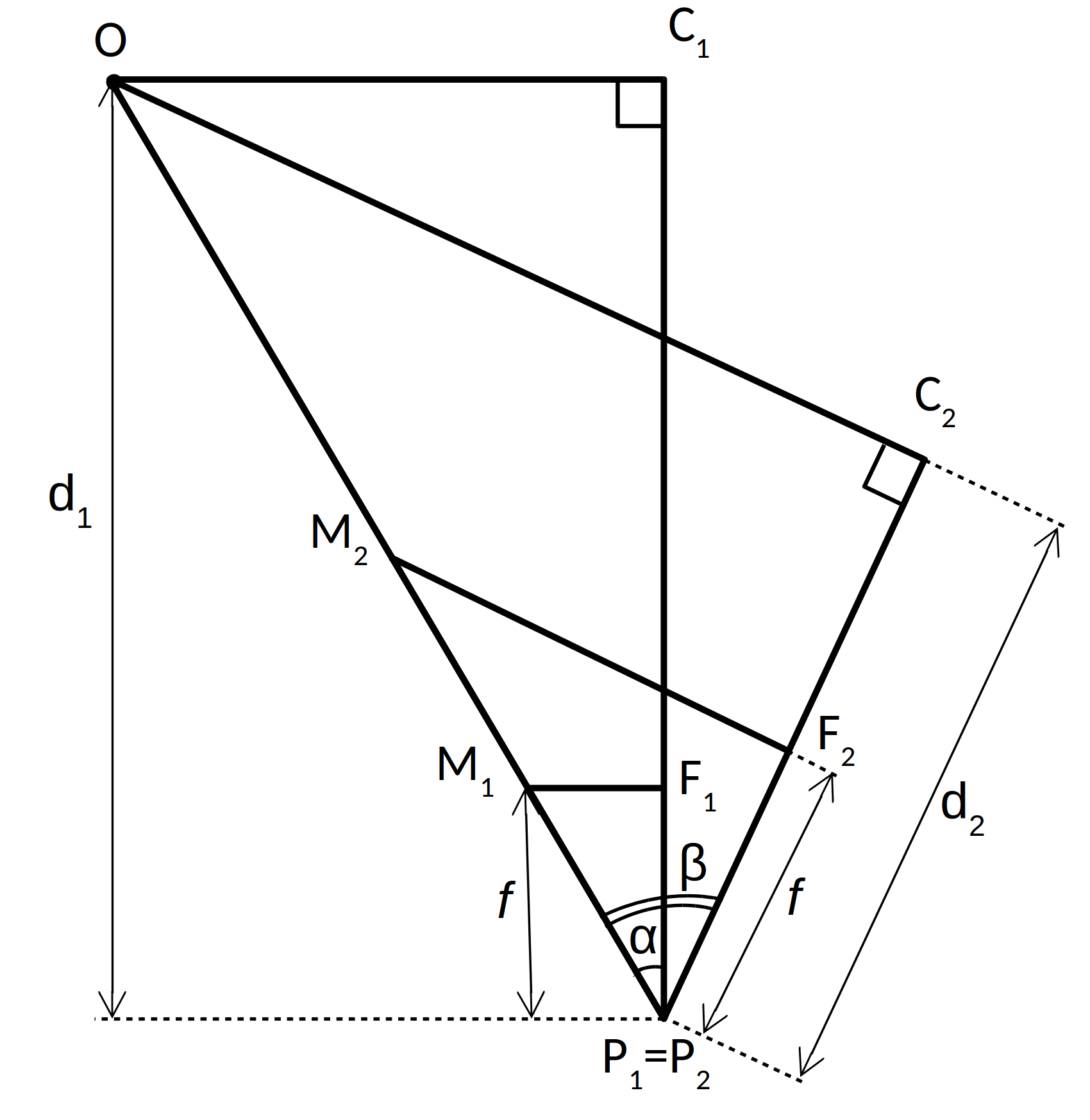}
        & \includegraphics[width=0.23\linewidth,scale=2]{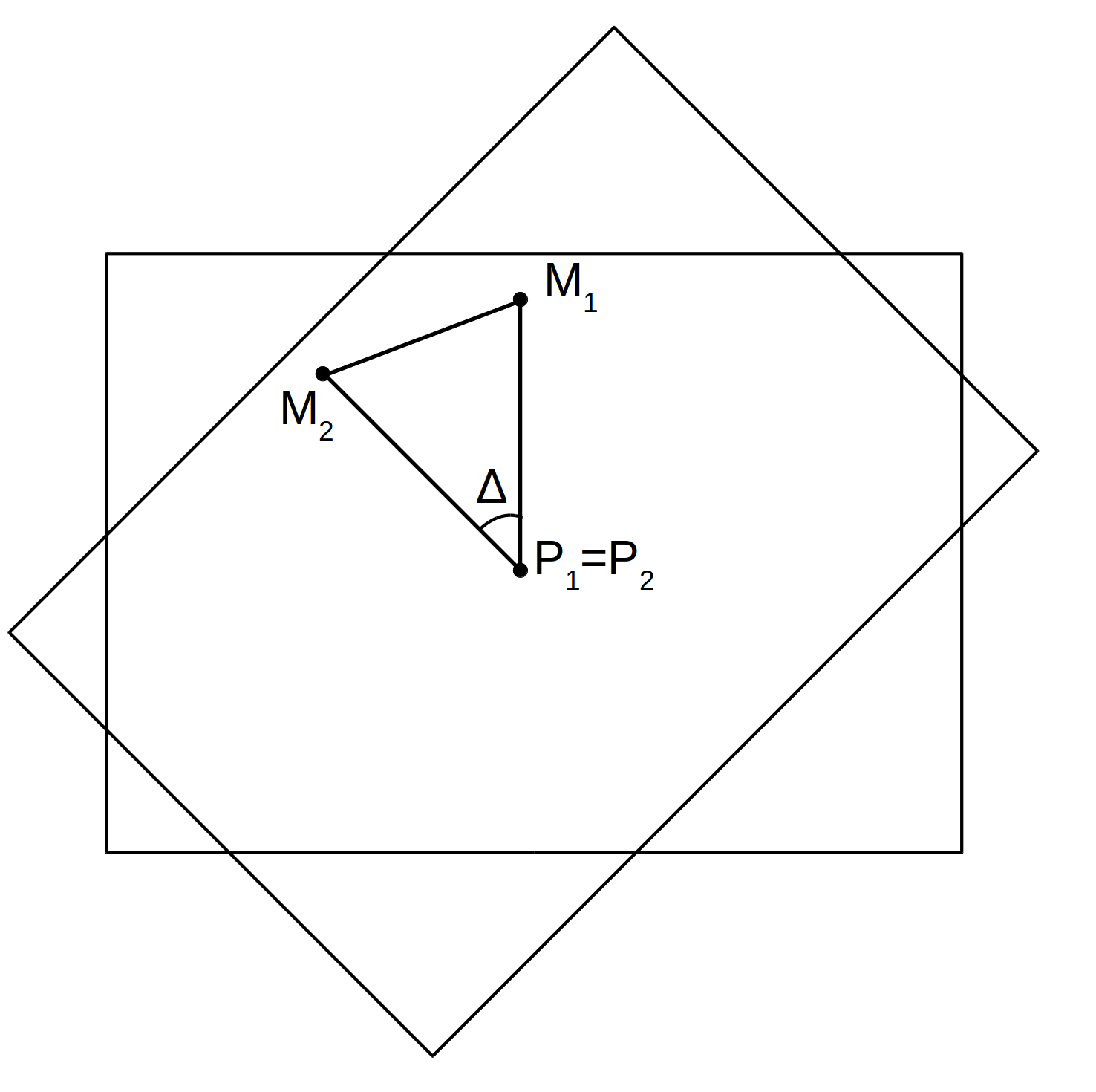} \\
        \hypertarget{figureA}{(A)} & \hypertarget{figureB}{(B)} & \hypertarget{figureC}{(C)} & \hypertarget{figureD}{(D)} \\
        Translations along & Translations along & Rotations with respect & Rotations with respect  \\
        x-axis or y-axis & z-axis & to x-axis or y-axis & to z-axis \\
    \end{tabular}
  \end{center}
\end{table*}

Let the camera motion be a combination of rotation and translation where translation is a 3-dimensional vector $(t_x, t_y, t_z)$ and rotation contains Euler angles $(r_x, r_y, r_z)$. 

Let $flow_x$,~$flow_y$ denote normalized OF maps along \mbox{x-axis} and \mbox{y-axis} respectively, with min/max values within interval $[-1, 1]$. Let $d$ denote depth. $grid_x, grid_y$ -- normalized size-invariant coordinate grids for pixels, with top left pixel having coordinates $(-1, -1)$ and bottom right pixel placing at $(1, 1)$. $f_x$,~$f_y$ denote focal lengths, \eg $f_x=0.5792$,~$f_y=1.9119$ for KITTI\cite{Geiger2012CVPR}. If subscript of the variables is missing, it is assumed that it can be either $x$~or~$y$ but identical for all variables in expression.

Then rotation and translation can be estimated as follows: 
\newline

    \begin{equation}
        \mathbf{t_x} = \frac{flow_x}{f_x} \times d
    \end{equation}
    \begin{equation} 
        \mathbf{t_y} = \frac{flow_y}{f_y} \times d 
    \end{equation}
    \begin{equation} 
        \mathbf{t^{(x)}_z} = -\frac{flow_x}{grid_x + flow_x} \times d
    \end{equation}
    \begin{equation} 
        \mathbf{t^{(y)}_z} = -\frac{flow_y}{grid_y + flow_y} \times d
    \end{equation}
    \begin{equation} 
        \mathbf{r_x} = \arctan \left( \frac{grid_x + flow_x}{f_x} \right) - \arctan \left(\frac{grid_x}{f_x}\right)
    \end{equation}
    \begin{equation} 
        \mathbf{r_y} = \arctan \left(\frac{grid_y + flow_y}{f_y}\right) - \arctan \left(\frac{grid_y}{f_y}\right)
    \end{equation}
    \begin{equation} 
        \mathbf{r_z} =  -\arccos \frac{u v } {|u||v|} \times sign( u \otimes v) 
    \end{equation} where
    \begin{equation} 
        u = \left(\frac{grid_x}{f_x} , \frac{grid_y}{f_y} \right)
    \end{equation}
    \begin{equation} 
        v = \left( \frac{grid_x + flow_x}{f_x} ,\frac{grid_y + flow_y}{f_y} \right) 
    \end{equation}
\\
All the proofs are provided below.

Hereinafter camera poses are denoted as $P_1, \ P_2$. 
Let also $C_1, \ C_2$ be principal points of a frustrum. For an arbitrary point $O$ from frustrum, $\angle O P_1 C_1$ is denoted as $\alpha$, $\angle O P_2 C_2$ as $\beta$. \\
\newline
\textbf{Translations along x-axis or y-axis} \hyperlink{figureA}{(A)}. For translations along x-axis or y-axis $t_x, \ t_y$ the proofs are identical, so they are presented in generalized form. 

So, the translation $t$ equals to $a - b$, where $a, \ b$ can be expressed using $\alpha, \ \beta$ as $a = \tan \alpha \times d, \ b = \tan \beta \times d$. 

Then OF can be used to find $\alpha, \ \beta$. It can be seen from the image that $\tan \alpha = \frac{M_1 F_1}{f}, \ \tan \beta = \frac{M_2 F_2}{f}$. Hence 
\begin{equation} 
    t = (\tan \alpha - \tan \beta) \times d = \left( \frac{M_1 F_1 - M_2 F_2}{f} \right) \times d
\end{equation}    
Finally, using that $M_2 F_2 = M_1 F_1 + flow$, we obtain 
\begin{equation} 
    t = \frac{flow}{f} \times d
\end{equation}
\newline
\textbf{Translations along z-axis} \hyperlink{figureB}{(B)}. Let us investigate shift along z-axis, or change of depth. Note that when there is no motion but along z-axis principal points of frustrum do coincide: $C = C_1 = C_2$.

The desired $t_z$ equals to $d_1 - d_2$. $d_2$ can be calculated as $\cot{\beta} \times O C$, and from triangle $O C_1 P_1$ we obtain $O C = \tan{\alpha} \times d_1$, thus $d_2 = \cot{\beta} \tan{\alpha}$. 

Looking at the triangles $M_1 F_1 P_1, \ M_2 F_2 P_2$  we can conclude that $\cot{\beta} = \frac{f}{M_2 F_2}, \ \tan \alpha = \frac{M_1 F_1}{f}$. 

Similarly to the previous case, $M_2 F_2 = M_1 F_1 + flow$, where $flow = flow_x$ or $flow_y$, depending on choice of x-axis or y-axis as reference coordinate axis for z-axis. \\
Then 
\begin{equation} 
    d_2 = \frac{f}{M_2 F_2} \frac{M_1 F_1}{f} \times d_1 = \frac{M_1 F_1}{M_2 F_2} \times d_1
\end{equation}
\begin{equation}
    \begin{split}
    t_z =& -d_1 - \frac{M_1 F_1}{M_2 F_2} \times d_1 -\left( 1 - \frac{M_1 F_1}{M_2 F_2} \right) \times d_1 = \\
    =& -\frac{M_2 F_2 - M_1 F_1}{M_2 F_2} \times d_1 -\frac{flow}{M_2 F_2} \times d_1 = \\
    =& -\frac{flow}{M_1 F_1 + flow} \times d_1
    \end{split}
\end{equation}
\newline
\textbf{Rotations with respect to x-axis or y-axis} \hyperlink{figureC}{(C)}. Similar to the translations $t_x, \ t_y$, we can write a single proof for rotations with respect to x-axis and y-axis, with $r$ standing for $r_x$ or $r_y$. The desired $r$ equals to $\beta - \alpha$. Here $\alpha = \arctan{\frac{M_1 F_1}{f}}, \ \beta = \arctan{\frac{M_2 F_2}{f}}$ and $M_2 F_2 = M_1 F_1 + flow$. \\
Thereby 
\begin{equation}
    r = \arctan{\frac{M_1 F_1 + flow}{f}} - \arctan{\frac{M_1 F_1}{f}}
\end{equation}
\newline
\textbf{Rotations with respect to z-axis} \hyperlink{figureD}{(D)}. Finally, for the case of rotation along z-axis, we can simply build point-to-pixel injection for two images.

Point $M_1$ is injected to pixel $u = (x, y)$, while $M_2$ is injected to $v = (x + flow_x, y + flow_y)$.

Now we can calculate cosine distance between these pixels and take $\arccos$ of it to obtain $r_z$: 
\begin{equation}
    r_z = -\arccos{\frac{u v}{ |u | |v| }}
\end{equation}
It is worth noting that inverted cosine function yields only non-negative values, so this formula gives absolute value of rotation angle. In order to distinguish clockwise and counter-clockwise rotation, we should consider the direction of rotation, which is determined by sign of the angle. Instead, we can calculate sign of sinus of the computed angle. In its turn, it is similar to the sign of cross product of $u, \ v$. So the final formula can be written as 
\begin{equation}
    r_z = -\arccos{\frac{u v}{ |u | |v| }} \times \sign{u \otimes v}
\end{equation}
\newline
\textbf{Decomposition results.} To make formulae demonstrative, let us examine a table containing motion maps of translations and rotations with respect to a single 6DoF component. It can be noticed that the pictures on diagonal are constant with each pixel value equal to the value of this 6DoF component shown in \autoref{fig:motion_maps}. In this simple example, it is sufficient to average diagonal maps to obtain 6DoF, so this problem can be solved by a simple deterministic algorithm. However, real motions are much more complex, thereby a more elaborated approaches such as the use of neural networks are needed. In practice we noticed that for realistic motions with one or two dominant components, the corresponding motion maps still are almost constant.

Moreover, taking a closer look on non-diagonal motion maps, one can deduce that in case of almost independent motions (\eg translation along x-axis and rotation along x-axis, or translation along y-axis) the corresponding features will be constant or nearly constant.

\begin{figure*}
    \begin{center}
        \includegraphics[width=1\linewidth,scale=2]{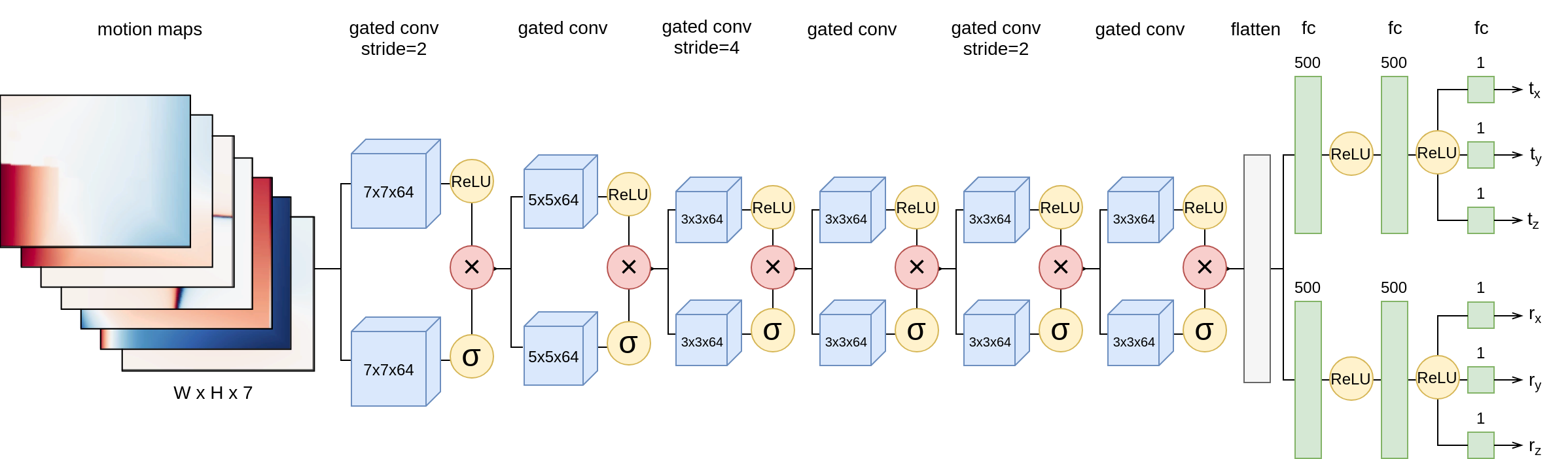}
        \caption{Network architecture. Inputs can be one of: OF, OF stacked with disparity or motion maps} \label{fig:new_net}
    \end{center}
\end{figure*}

\section{Egomotion estimation network}
The proposed network architecture is shown in  \autoref{fig:new_net}. It takes 7 stacked motion maps as input and outputs 6DoF. The network itself consists of a feature extractor followed by a multi-layer regressor. The output of the feature extractor is reshaped into a single-dimensional vector and fed into the regressor to obtain 6DoF.

The feature extractor is composed from 6 convolutional layers placed sequentially. In our work, we used gated convolutions \cite{lei2015semi} that show better results then regular ones. Each convolution has 64 channels. Kernel sizes of gated convolutions are [7x7,~5x5,~3x3,~3x3,~3x3,~3x3] with strides [2,~1,~4,~1,~2,~1] respectively. The regressor consists of two separate branches for translation and rotation respectively, as such split proved to perform better. Every branch consists of three fully-connected layers.

As this architecture is quite simple, it is crucial to preserve as much spatial information as possible. Therefore strided convolutions are used instead of max-pooling.

\section{Experiments}

\subsection{Baselines}
We implemented several other network architectures for comprehensive comparison. Training protocol for all baselines follows \ref{subsec:training_procedure} unless otherwise specified. OF and depth inputs are provided by PWC and Struct2Depth networks (see \ref{subsec:backbone}).\\
\newline
\textbf{LS-VO\cite{costante2018ls}.} In given approach regression network is supplied with an autoencoder part, transforming OF into a latent representation and then reconstructing it back. The difference between our implementation and the authors implementation is that we do not fine-tune OF estimation network. Furthermore, input sizes differ: (96px,~300px) is listed in original paper, while in our case PWC-net outputs OF with size (96px,~320px), which makes cropping layer from original architecture irrelevant. Surprisingly, our studies showed that removing decoder branch leads to slightly better results.\\
\newline
\textbf{L-VO(3D-Flow)\cite{zhao2018learning}.} The authors did not provide implementation of their proposed method. Our implementation followed the paper, but we did not implement the prediction of Gaussian distribution parameters for translations as well as corresponding loss function. Yet, such implementation helps to compare provided architecture without specified additional feature. Since the training procedure (\ref{subsec:training_procedure}) did not converge due to large initial learning rate, we used the training procedure specified in the original paper. Our experiments showed that removing depth-flow branch leads to slightly better and more stable results.\\
\newline
\textbf{RTN\cite{Lv18eccv}.} This network attracted our attention by using coordinate grids as one of the inputs. Such trick seems quite reasonable, since pixel coordinates information is essential for ego-motion estimation (see \ref{subsec:calculation_of_motion_maps}). Authors have submitted network code in PyTorch and we re-implemented it in Keras. In our implementation, the network lacks a decoder for rigidity segmentation since datasets used in our study do not contain corresponding ground truth data.

\subsection{Implementation details}
\noindent \textbf{OF and depth estimation networks.}\label{subsec:backbone}
We obtain dense OF from pairs of RGB images using PWC-Net\cite{sun2018pwc}, which is the common choice for flow estimation task. PWC-Net was pretrained on flyingthings3D\cite{MIFDB16} dataset, network realization and weights are borrowed from official repository\footnote{\url{https://github.com/NVlabs/PWC-Net}}.

If ground truth depth is not present, we estimate depth from RGB inputs with recently introduced Struct2Depth\cite{casser2019struct2depth} -- unsupervised method for depth and ego-motion estimation. In our work, we use model from official repository\footnote{\url{https://github.com/tensorflow/models/tree/master/research/struct2depth}} that was pretrained on KITTI dataset. The authors provide the trained models along with implementation.\\
\newline
\textbf{Training procedure.}\label{subsec:training_procedure}
Our network is trained to minimize MAE for KITTI and Huber loss for DISCOMAN benchmark. Surprisingly, opting for L1 loss leads to less stable results comparing to L2, but this improves performance significantly. 

\begin{figure*}[t]
    \begin{center}
        \includegraphics[width=0.85\linewidth,scale=2]{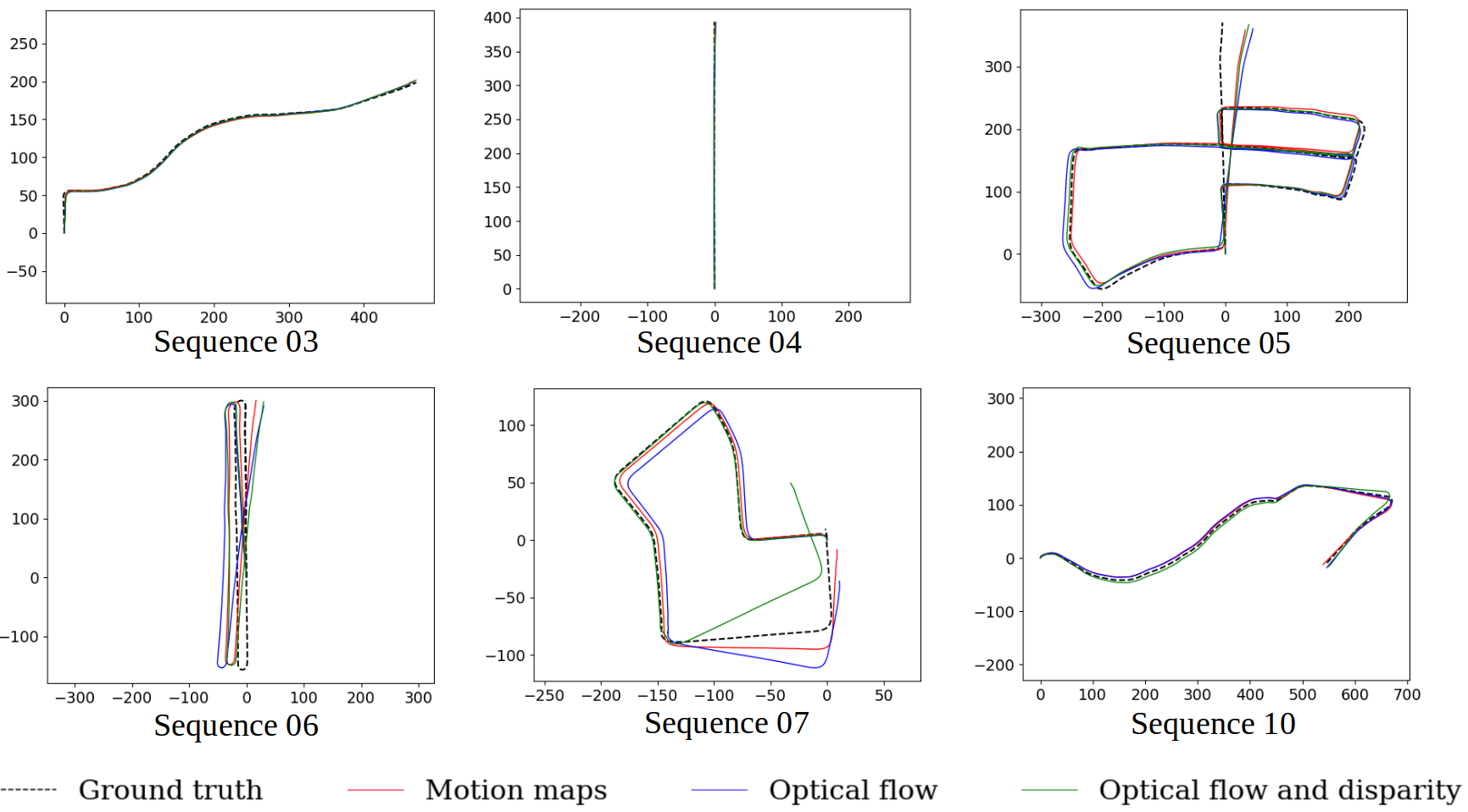}
        \caption{Predictions of our approach using different inputs on KITTI}
    \end{center}
\end{figure*}

\begin{table*}[t]
  \begin{center}
    \caption{Comparison with published results on KITTI}
    \label{tab:table1}
    \begin{tabular}{|l|c|c|c|c|c|c|c|c|c|c|c|c|c|c|}
        \hline
        \multirow{3}{*}{\textbf{Method}} & \multicolumn{12}{c|}{\textbf{Sequence}} & \multicolumn{2}{c|}{\multirow{2}{*}{Average}} \\
        \cline{2-13}
        & \multicolumn{2}{c|}{03} & \multicolumn{2}{c|}{04} & \multicolumn{2}{c|}{05} & \multicolumn{2}{c|}{06} & \multicolumn{2}{c|}{07} & \multicolumn{2}{c|}{10} & \multicolumn{2}{c|}{} \\
        \cline{2-15}
        & $t_{rel}$ & $r_{rel}$ & $t_{rel}$ & $r_{rel}$ & $t_{rel}$ & $r_{rel}$ & $t_{rel}$ & $r_{rel}$ & $t_{rel}$ & $r_{rel}$ & $t_{rel}$ & $r_{rel}$ & $t_{rel}$ & $r_{rel}$ \\
        \hline
        2D-Flow\cite{zhao2018learning} & 3.35 & 1.62 & 4.15 & 2.53 & 2.49 & 1.19 & 3.19 & 1.54 & 17.2 & 10.4 & 7.24 & 3.06 & 6.27 & 3.39 \\
        3D-Flow\cite{zhao2018learning} & 3.18 & 1.31 & 2.04 & 0.81 & 2.59 & 0.99 & 1.39 & 0.95 & 2.81 & 2.54 & 4.38 & 3.12 & 2.73 & 1.62 \\
        VISO-S\cite{Geiger2011StereoScanD3} & 1.71 & 1.12 & 1.54	& 0.84 & 2.36 & 1.20 & 1.47	& 0.87 & 2.37 & 1.78 & 1.51	& 1.15 & \textbf{1.83} & 1.16 \\
        VISO2-S\cite{Geiger2011StereoScanD3}	& 3.21 & 3.25 & 2.12 & 2.12	& 1.53 & 1.60 & 1.48 & 1.58	& 1.85 & 1.91 & 1.17 & 1.30 & 1.89 & 1.96 \\
        UnDeepVO\cite{Li2018UnDeepVOMV} & 5.00 & 6.17 & 5.49 & 2.13 & 3.40 & 1.50 & 6.20 & 1.98 & 3.15 & 2.48 & 10.63 & 4.65 & 5.65 & 3.15 \\
        DeepVO\cite{wang2017deepvo} & 8.49 & 6.89 & 7.19	& 6.97 & 2.62 & 3.61 & 5.42	& 5.82 & 3.91 & 4.60 & 8.11	& 8.83 & 5.96 & 6.12 \\
        ESP-VO\cite{wang2018end} & 6.72 & 6.46 & 6.33 & 6.08	& 3.35 & 4.93 & 7.24 & 7.29 & 3.52 & 5.02 & 9.77 & 10.2	& 6.15& 6.63 \\
        SRNN\_{channel}\cite{Xue2018GuidedFS} & 5.44 & 3.32 & 2.91 & 1.30 & 3.27 & 1.62 & 8.50 & 2.74 & 3.37 & 2.25 & 6.32 & 2.33 & 4.80 & 2.26 \\
        LS-VO\cite{costante2018ls}& 5.30 & 1.53 & 0.78 & 0.42 & 2.18 & 0.91 & 2.93 & 1.14 & 10.20 & 5.53 & 3.71 & 1.26 & 4.18 & 1.80 \\
        \hline
        Ours &2.38 & 0.93 &0.78 &0.27 &2.36 &0.87 &2.91 &1.14 &3.51 &1.44 &3.31 &1.11 &2.54 &\textbf{0.96} \\
        \hline
    \end{tabular}
  \end{center}
\end{table*}

\begin{table*}[t]
  \begin{center}
    \caption{Comparison with implemented baselines on KITTI}
    \label{tab:table2}
    \begin{tabular}{|l|l|l|c|c|c|}
    \hline
    Inputs & Model & Loss & ATE & RMSE\textsubscript{t} & RMSE\textsubscript{r} \\ 
    \hline
    OF, disparity & 3D-Flow\cite{zhao2018learning} & MSE & 15.14 $\pm$ 1.92 & 6.40 $\pm$ 0.70 & 3.05 $\pm$  0.31 \\
    RGB-D & RTN\cite{Lv18eccv} & MAE & 	
72.12 $\pm$ 6.13 & 44.85 $\pm$ 5.52 & 18.17 $\pm$ 2.55 \\
    OF & LS-VO\cite{costante2018ls} & MAE & 6.25 $\pm$ 0.77 & 3.43 $\pm$ 0.34 & 1.42 $\pm$ 0.10 \\
    \hline
    OF & Ours & MAE & 6.37 $\pm$ 1.83 & 2.88 $\pm$ 0.81 & 1.28 $\pm$ 0.30 \\
    OF, disparity & Ours & MAE & 5.39 $\pm$ 0.89 & \textbf{2.53 $\pm$ 0.30} & 1.24 $\pm$ 0.19 \\ 
    Motion maps & Ours & MAE & \textbf{5.32 $\pm$ 1.16} & 2.54 $\pm$ 0.38 & \textbf{0.96 $\pm$ 0.14} \\
    \hline
    \end{tabular}
  \end{center}
\end{table*}

The network is trained from scratch using Adam optimization with amsgrad option switched on. The batch size is set to 128, the momentum is fixed to (0.9, 0.999). Initial learning rate is 0.001 and is multiplied by 0.5 if validation loss does not improve for 10 epochs. Training process is terminated when learning rate becomes negligibly small -- we used $10^{-5}$ as a threshold. Under this conditions, models were typically trained for about 100 epochs.

In several papers on trainable visual odometry \cite{costante2018ls, Lv18eccv, wang2017deepvo, zhao2018learning, zhou2018deeptam} different weights are used for translation loss and rotation loss due to several reasons. Firstly, even small rotation errors may have a crucial impact on the shape of trajectory, so precise estimation or Euler angles is more important comparing to translations. Moreover, in different datasets translations scales vary. Thus, it is useful to adjust loss weights addressing translation and rotation components independently. We multiply loss for rotation components by 50, as it was proposed in \cite{costante2018ls}.

\subsection{Datasets}
We concentrated our efforts on datasets containing "almost flat" egomotion along a 2D surface, namely KITTI Visual Odometry benchmark\cite{Geiger2012CVPR} collected via a camera placed on top of a vehicle, and a synthetic DISCOMAN dataset, where camera motion imitates behaviour of a home robot exploring indoor scenes. By virtue of the moving agent, both of the datasets contain trajectories where vertical movements are eliminated or present as minor fluctuations. \\
\newline
\textbf{KITTI VO}\cite{Geiger2012CVPR} is a basic benchmark for learnable odometry algorithms. It consists of 22 sequences saved in PNG format. Sequences 00-10 provide the sensor data with the accurate ground truth ($<$10cm) from a GPS/IMU system, while sequences 11-21 only provide the raw sensor data. While there are several way of how to split sequences train/val, we chose to use sequences [00,~01,~02,~08] as train and  [03,~04,~05,~06,~07,~10] as validation. Such split allowed us to compare results with larger number of published approaches. \\
\newline
\textbf{DISCOMAN} (in submission, attached). Recently launched Dataset of Indoor SCenes for Odometry, Mapping And Navigation provides variety of sequences of synthetic indoor images supplied with ground truth poses, precise depth maps and OF. To speed up experiments, we trained our networks on 51 randomly selected sequences from the train set while 17 randomly selected sequences from validation set where used for testing. Since camera motions is relatively small in this dataset, we skipped every second frame. In experiments with motion maps as inputs we used only $t_x, t_y, r_x, r_y$ components, since other motion maps appear to be constant due to synthetic nature of the data.

\begin{figure*}[!t]
    \begin{center}
        \includegraphics[width=0.85\linewidth,scale=2]{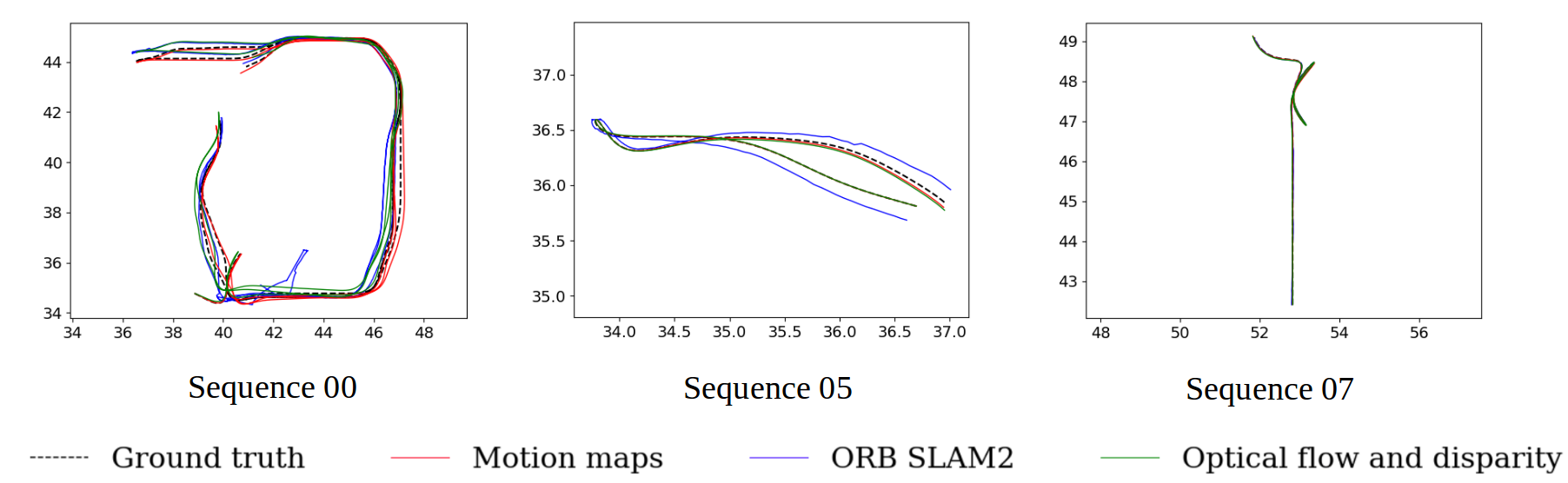}
        \caption{Predictions of ORB SLAM2 and our approach on DISCOMAN trajectories}
    \end{center}
\end{figure*}

\begin{table*}[t]
    \begin{center}
    \caption{Comparison with ORB SLAM2 on DISCOMAN}
    \label{tab:table3}
    \begin{tabular}{|l|l|l|c|c|c|c|c|c|}
    \hline
    \multirow{2}{*}{Method} & \multicolumn{4}{c|}{Easy for ORB SLAM2 trajectories} & \multicolumn{4}{c|}{All trajectories} \\ 
    \cline{2-9}
    & Success rate & ATE & RPE\textsubscript{t} & RPE\textsubscript{r} & Success rate & ATE & RPE\textsubscript{t} & RPE\textsubscript{r} \\ \hline
    ORB SLAM2 (RGB-D) & 100\% & 0.039 & 0.053 & 0.916 & 53\% & - & \multicolumn{1}{c|}{-} & \multicolumn{1}{c|}{-} \\ \hline
    Ours (motion maps) & 100\% & 0.047 & 0.158 & 1.556 & 100\% & 0.057 & 0.218 & 1.352 \\ \hline
    \end{tabular}
    \end{center}
\end{table*}

\begin{table*}[!t]
    \begin{center}
    \caption{Comparison with implemented baselines on DISCOMAN}
    \label{tab:table4}
    \begin{tabular}{|l|l|l|c|c|c|}
    \hline
    Inputs & Model & Loss & ATE & RPE\textsubscript{t} & RPE\textsubscript{r} \\ 
    \hline
    OF, disparity & 3D-Flow\cite{zhao2018learning} & MSE & 0.250 $\pm$ 0.040 & 1.085 $\pm$ 0.217 & 7.548 $\pm$ 1.475 \\ 
    RGB-D & RTN\cite{Lv18eccv} & MSE & 2.098 $\pm$ 0.050 & 9.896 $\pm$ 0.807 & 79.145 $\pm$ 9.465 \\ 
    OF & LS-VO\cite{costante2018ls} & Huber & 0.077 $\pm$ 0.007 & 0.274 $\pm$ 0.022 & 1.511 $\pm$ 0.119 \\ 
    \hline
    OF & Ours & Huber & 0.066 $\pm$ 0.010 & 0.230 $\pm$ 0.046 & 1.480 $\pm$ 0.330 \\ 
    OF, disparity & Ours & Huber & 0.088 $\pm$ 0.014 & 0.345 $\pm$ 0.074 & 2.138 $\pm$ 0.491 \\ 
    Motion maps & Ours & Huber & \textbf{0.057 $\pm$ 0.001} & \textbf{0.218 $\pm$ 0.015} & \textbf{1.352 $\pm$ 0.150} \\ 
    \hline
    \end{tabular}
\end{center}
\end{table*}

\subsection{Metrics} 
We evaluate visual odometry methods with the KITTI evaluation metrics, computing the average translation and rotation errors for the set of sub-sequences. As there are only few sequences in KITTI, many researches preferred to compare results sequence-wise, as more detailed examination helped to reveal strengths and weaknesses of different methods. Final values were obtained by simply averaging metrics for every single trajectory. Following \cite{Sturm2012ABF}, translation and rotation errors were calculated as Root Mean Squared Error for all possible sub-sequences of length (100,~\dots,~800) meters, and usually referred as RMSE.

Alternative protocol is implemented in KITTI devkit \footnote{\url{https://github.com/alexkreimer/odometry/devkit}} that introduces translation and rotation Relative Pose Error, referred as RPE. The difference between RMSE and RPE is twofold. Firstly, to calculate rotation RPE, not squared but absolute rotation errors are averaged. This accumulation strategy causes rotation RPE to be more robust comparing to rotation RMSE.

The second difference is the averaging strategy. In calculation of RMSE, all the sequences and all the sub-sequences of each sequence contribute equally to the value of a final metric. Therefore, sequences with fewer frames are considered to be, counter-intuitively, as important as the longest ones.
To obtain RPE, translation and rotation errors are averaged for all sub-sequences of all sequences. In contrast to RMSE, there is no intermediate averaging by sequence, so contribution of each sequence is proportional to the number of its sub-sequences.

In our studies, we found out that RPE is more stable in terms of standard deviation to average ratio. However, to compare our method with a wider range of approaches we provide RMSE values for KITTI dataset. For DISCOMAN, we opted for RPE, as there are no previous works that might restrict our choice of metrics. 

In addition, to give a full picture we also provide values of Average Translation Error, or ATE.

The results of networks trained with different random initialization vary significantly. Thus, in order to obtain most fair comparison we run all experiments for 5 times with different random seeds. The metrics reported are mean and standard deviation of execution results.

\subsection{Results}
On KITTI our approach outperforms monocular methods in terms of both rotations and translations (\autoref{tab:table1}). Moreover, as indicated by the impressive values of RMSE\textsubscript{r}, it predicts rotations more accurately in comparison to both learnable and classical VISO-S and VISO2-S methods. 

In another set of experiments on KITTI (\autoref{tab:table2}) we analyze how the choice of inputs type affects model performance. Through our evaluation, we show that utilizing motion maps leads to a significant increase of RMSE\textsubscript{r} accompanied with competitive results in terms of other metrics.  

For DISCOMAN, we evaluate our method against trainable and classical methods, the latter represented by ORB SLAM2 (\autoref{tab:table3}). Comparison with ORB SLAM2 is not completely fair since this method exploit bundle adjustment. However, our approach yields satisfactory results while demonstrating much higher robustness.    

Moreover, exploiting motion maps on DISCOMAN caused an improvement of ATE, RMSE\textsubscript{r} and RMSE\textsubscript{t}, with all the results being significantly more stable, according to standard deviation (\autoref{tab:table4}).

\section{Conclusions}

We introduced a novel method of transforming OF and depth maps into several \textit{motion maps}, each one corresponding to a single degree of freedom. The calculation of motion maps is based on projective geometry and requires minimum computation. Through extensive evaluation, we showed that using motion maps combined with our proposed network architecture results in an improvement on KITTI and DISCOMAN benchmarks.

\newpage

{\small
\bibliographystyle{ieee}
\bibliography{references}

\begin{thebibliography}{10}\itemsep=-1pt

\bibitem{Ambarella}
Ambarella cvflow technology overview.
\newblock \url{https://www.ambarella.com/technology/technology-overview}.
\newblock Accessed: 2018-10-30.

\bibitem{nvidia_of_sdk}
Nvidia optical flow sdk.

\bibitem{almalioglu2018ganvo}
Y.~Almalioglu, M.~R.~U. Saputra, P.~P. de~Gusmao, A.~Markham, and N.~Trigoni.
\newblock Ganvo: Unsupervised deep monocular visual odometry and depth
  estimation with generative adversarial networks.
\newblock {\em arXiv preprint arXiv:1809.05786}, 2018.

\bibitem{casser2019struct2depth}
V.~Casser, S.~Pirk, R.~Mahjourian, and A.~Angelova.
\newblock Unsupervised learning of depth and ego-motion: A structured approach.
\newblock In {\em Thirty-Third AAAI Conference on Artificial Intelligence
  (AAAI-19)}, 2019.

\bibitem{costante2018ls}
G.~Costante and T.~A. Ciarfuglia.
\newblock Ls-vo: Learning dense optical subspace for robust visual odometry
  estimation.
\newblock {\em IEEE Robotics and Automation Letters}, 3(3):1735--1742, 2018.

\bibitem{dharmasiri2018eng}
T.~Dharmasiri, A.~Spek, and T.~Drummond.
\newblock Eng: End-to-end neural geometry for robust depth and pose estimation
  using cnns.
\newblock {\em arXiv preprint arXiv:1807.05705}, 2018.

\bibitem{dosovitskiy2015flownet}
A.~Dosovitskiy, P.~Fischer, E.~Ilg, P.~Hausser, C.~Hazirbas, V.~Golkov, P.~Van
  Der~Smagt, D.~Cremers, and T.~Brox.
\newblock Flownet: Learning optical flow with convolutional networks.
\newblock In {\em Proceedings of the IEEE International Conference on Computer
  Vision}, pages 2758--2766, 2015.

\bibitem{engel2018direct}
J.~Engel, V.~Koltun, and D.~Cremers.
\newblock Direct sparse odometry.
\newblock {\em IEEE transactions on pattern analysis and machine intelligence},
  40(3):611--625, 2018.

\bibitem{forster2014svo}
C.~Forster, M.~Pizzoli, and D.~Scaramuzza.
\newblock Svo: Fast semi-direct monocular visual odometry.
\newblock In {\em Robotics and Automation (ICRA), 2014 IEEE International
  Conference on}, pages 15--22. IEEE, 2014.

\bibitem{fu2018deep}
H.~Fu, M.~Gong, C.~Wang, K.~Batmanghelich, and D.~Tao.
\newblock Deep ordinal regression network for monocular depth estimation.
\newblock In {\em Proceedings of the IEEE Conference on Computer Vision and
  Pattern Recognition}, pages 2002--2011, 2018.

\bibitem{Geiger2012CVPR}
A.~Geiger, P.~Lenz, and R.~Urtasun.
\newblock Are we ready for autonomous driving? the kitti vision benchmark
  suite.
\newblock In {\em Conference on Computer Vision and Pattern Recognition
  (CVPR)}, 2012.

\bibitem{Geiger2011StereoScanD3}
A.~Geiger, J.~Ziegler, and C.~Stiller.
\newblock Stereoscan: Dense 3d reconstruction in real-time.
\newblock {\em 2011 IEEE Intelligent Vehicles Symposium (IV)}, pages 963--968,
  2011.

\bibitem{ilg2017flownet}
E.~Ilg, N.~Mayer, T.~Saikia, M.~Keuper, A.~Dosovitskiy, and T.~Brox.
\newblock Flownet 2.0: Evolution of optical flow estimation with deep networks.
\newblock In {\em IEEE conference on computer vision and pattern recognition
  (CVPR)}, volume~2, page~6, 2017.

\bibitem{kerl2013robust}
C.~Kerl, J.~Sturm, and D.~Cremers.
\newblock Robust odometry estimation for rgb-d cameras.
\newblock In {\em Robotics and Automation (ICRA), 2013 IEEE International
  Conference on}, pages 3748--3754. IEEE, 2013.

\bibitem{lei2015semi}
T.~Lei, H.~Joshi, R.~Barzilay, T.~Jaakkola, K.~Tymoshenko, A.~Moschitti, and
  L.~Marquez.
\newblock Semi-supervised question retrieval with gated convolutions.
\newblock {\em arXiv preprint arXiv:1512.05726}, 2015.

\bibitem{Li2018UnDeepVOMV}
R.~Li, S.~Wang, Z.~Long, and D.~Gu.
\newblock Undeepvo: Monocular visual odometry through unsupervised deep
  learning.
\newblock {\em 2018 IEEE International Conference on Robotics and Automation
  (ICRA)}, pages 7286--7291, 2018.

\bibitem{liang2018learning}
Z.~Liang, Y.~Feng, Y.~Chen, and L.~Zhang.
\newblock Learning for disparity estimation through feature constancy.
\newblock In {\em Proceedings of the IEEE Conference on Computer Vision and
  Pattern Recognition}, pages 2811--2820, 2018.

\bibitem{luo2016efficient}
W.~Luo, A.~G. Schwing, and R.~Urtasun.
\newblock Efficient deep learning for stereo matching.
\newblock In {\em Proceedings of the IEEE Conference on Computer Vision and
  Pattern Recognition}, pages 5695--5703, 2016.

\bibitem{Lv18eccv}
Z.~Lv, K.~Kim, A.~Troccoli, D.~Sun, J.~Rehg, and J.~Kautz.
\newblock Learning rigidity in dynamic scenes with a moving camera for 3d
  motion field estimation.
\newblock In {\em ECCV}, 2018.

\bibitem{MIFDB16}
N.~Mayer, E.~Ilg, P.~H{\"a}usser, P.~Fischer, D.~Cremers, A.~Dosovitskiy, and
  T.~Brox.
\newblock A large dataset to train convolutional networks for disparity,
  optical flow, and scene flow estimation.
\newblock In {\em IEEE International Conference on Computer Vision and Pattern
  Recognition (CVPR)}, 2016.
\newblock arXiv:1512.02134.

\bibitem{mur2017orb}
R.~Mur-Artal and J.~D. Tard{\'o}s.
\newblock Orb-slam2: An open-source slam system for monocular, stereo, and
  rgb-d cameras.
\newblock {\em IEEE Transactions on Robotics}, 33(5):1255--1262, 2017.

\bibitem{steinbrucker2011real}
F.~Steinbr{\"u}cker, J.~Sturm, and D.~Cremers.
\newblock Real-time visual odometry from dense rgb-d images.
\newblock In {\em Computer Vision Workshops (ICCV Workshops), 2011 IEEE
  International Conference on}, pages 719--722. IEEE, 2011.

\bibitem{Sturm2012ABF}
J.~Sturm, N.~Engelhard, F.~Endres, W.~Burgard, and D.~Cremers.
\newblock A benchmark for the evaluation of rgb-d slam systems.
\newblock {\em 2012 IEEE/RSJ International Conference on Intelligent Robots and
  Systems}, pages 573--580, 2012.

\bibitem{sun2018pwc}
D.~Sun, X.~Yang, M.-Y. Liu, and J.~Kautz.
\newblock Pwc-net: Cnns for optical flow using pyramid, warping, and cost
  volume.
\newblock In {\em Proceedings of the IEEE Conference on Computer Vision and
  Pattern Recognition}, pages 8934--8943, 2018.

\bibitem{ummenhofer2017demon}
B.~Ummenhofer, H.~Zhou, J.~Uhrig, N.~Mayer, E.~Ilg, A.~Dosovitskiy, and
  T.~Brox.
\newblock Demon: Depth and motion network for learning monocular stereo.
\newblock In {\em IEEE Conference on computer vision and pattern recognition
  (CVPR)}, volume~5, page~6, 2017.

\bibitem{wang2017deepvo}
S.~Wang, R.~Clark, H.~Wen, and N.~Trigoni.
\newblock Deepvo: Towards end-to-end visual odometry with deep recurrent
  convolutional neural networks.
\newblock In {\em Robotics and Automation (ICRA), 2017 IEEE International
  Conference on}, pages 2043--2050. IEEE, 2017.

\bibitem{wang2018end}
S.~Wang, R.~Clark, H.~Wen, and N.~Trigoni.
\newblock End-to-end, sequence-to-sequence probabilistic visual odometry
  through deep neural networks.
\newblock {\em The International Journal of Robotics Research},
  37(4-5):513--542, 2018.

\bibitem{icra_2019_fastdepth}
{Wofk, Diana and Ma, Fangchang and Yang, Tien-Ju and Karaman, Sertac and Sze,
  Vivienne}.
\newblock {FastDepth: Fast Monocular Depth Estimation on Embedded Systems}.
\newblock In {\em {IEEE International Conference on Robotics and Automation
  (ICRA)}}, {2019}.

\bibitem{Xue2018GuidedFS}
F.~Xue, Q.~Wang, X.~Wang, W.~Dong, J.~Wang, and H.~Zha.
\newblock Guided feature selection for deep visual odometry.
\newblock {\em CoRR}, abs/1811.09935, 2018.

\bibitem{zhao2018learning}
C.~Zhao, L.~Sun, P.~Purkait, T.~Duckett, and R.~Stolkin.
\newblock Learning monocular visual odometry with dense 3d mapping from dense
  3d flow.
\newblock {\em Intelligent Robots and Systems (IROS), 2018 International
  Conference on}, 2018.

\bibitem{zhou2018deeptam}
H.~Zhou, B.~Ummenhofer, and T.~Brox.
\newblock Deeptam: Deep tracking and mapping.
\newblock In {\em European Conference on Computer Vision (ECCV)}, 2018.

\end{thebibliography}
}

\end{document}